# ASSESSING THE VIABILITY OF UNSUPERVISED LEARNING WITH AUTOENCODERS FOR PREDICTIVE MAINTENANCE IN HELICOPTER ENGINES

**P. Sánchez, K. Reyes, B. Radu, E. Fernández**

Department of Computer Science, University of Alcalá, Spain

## ABSTRACT

Unplanned engine failures in helicopters can lead to severe operational disruptions, safety hazards, and costly repairs. To mitigate these risks, this study compares two predictive maintenance strategies for helicopter engines: a supervised classification pipeline and an unsupervised anomaly detection approach based on autoencoders (AEs). The supervised method relies on labelled examples of both normal and faulty behaviour, while the unsupervised approach learns a model of normal operation using only healthy engine data, flagging deviations as potential faults. Both methods are evaluated on a real-world dataset comprising labelled snapshots of helicopter engine telemetry. While supervised models demonstrate strong performance when annotated failures are available, the AE achieves effective detection without requiring fault labels, making it particularly well suited for settings where failure data are scarce or incomplete. The comparison highlights the practical trade-offs between accuracy, data availability, and deployment feasibility, and underscores the potential of unsupervised learning as a viable solution for early fault detection in aerospace applications.

## I. INTRODUCTION

Helicopter engines are complex and safety-critical systems where an unexpected in-flight failure can be catastrophic. Even on the ground, unscheduled engine downtime can disrupt operations and incur steep costs for repairs and mission delays. To mitigate these risks, the aviation industry has embraced predictive maintenance, aiming to detect incipient engine faults before they escalate. Conventional predictive maintenance approaches often rely on supervised learning models that predict metrics like Remaining Useful Life (RUL) or classify fault states (Carvalho et al., 2019; Lei et al., 2018). However, supervised methods require substantial historical data covering many failure examples for training. In practice, such run-to-failure datasets for helicopter engines are extremely limited: maintenance crews typically service or replace components long before a total failure occurs, meaning fully labelled degradation trajectories are scarce.

In such data-scarce settings, unsupervised anomaly detection offers a pragmatic alternative. Rather than learning to predict time-to-failure, an unsupervised algorithm learns the normal patterns of engine operation and raises an alert whenever new sensor readings deviate significantly from this learned baseline. Among unsupervised approaches, neural network AEs have gained popularity for condition monitoring (Fathi et al., 2021). An AE is trained to compress and reconstruct its input; when trained on only healthy engine data, it will reproduce normal behaviour with low error, while failing to accurately reconstruct data containing a fault. In effect, the reconstruction error becomes an anomaly score: if an incoming sensor observation yields a large error, it is likely outside the distribution of normal operating data and thus indicative of a potential fault. This approach obviates the need for any fault labels, which is a major advantage when only abundant recordings of normal operation are available.

At the same time, if labelled examples of past faults are available, supervised learning can directly learn decision boundaries between normal and faulty states and potentially achieve higher accuracy (Susto et al., 2014). The question we investigate is how a state-of-the-art unsupervised method compares to traditional supervised classifiers for early fault detection in helicopter engines, and whether the unsupervised AE approach is feasible in scenarios where fault labels are sparse. To that end, we perform a head-to-head comparison using a dataset of helicopter engine sensor data. We evaluate multiple supervised classifiers (including ensemble methods and neural networks) against an AE-based anomaly detector. The comparison sheds light on the trade-offs between the two paradigms in terms of detection performance and practical deployment considerations. In particular, we highlight how the availability of labelled fault data impacts each method's performance and discuss the implications for real-world engine health monitoring.

In addition to classical supervised classifiers, many predictive maintenance studies have highlighted the potential of a broader family of supervised algorithms, including Logistic Regression, Linear and Quadratic Discriminant Analysis, Decision Trees, Random Forests, k-Nearest Neighbours, Support Vector Machines, and Artificial Neural Networks (Dreisetl et al., 2002; Anderson, 1958; Breiman, 1984; Hastie et al., 2017), which provide complementary modelling assumptions and robustness profiles. These models differ in their sensitivity to outliers, ability to capture non-linear relationships, interpretability, and computational efficiency, making them particularly relevant when evaluating multiple supervised baselines for helicopter engine data.

## II. DATA SCIENCE METHODOLOGY

Data Science and Artificial Intelligence, unlike more established fields such as Software Engineering, have traditionally lacked comprehensive methodological frameworks capable of covering the full spectrum of activities involved in the design, development, and deployment of AI-based solutions. Early attempts to formalize these processes did not appear until the late 1980s, largely as a result of combining principles from Software Engineering with advances in Artificial Intelligence research. However, it was not until the early 2000s that a broadly accepted methodological framework specifically targeting Data Mining was introduced: the Cross-Industry Standard Process for Data Mining (CRISP-DM) (Shearer, 2000). Since its publication, CRISP-DM has been widely adopted by both academic and industrial communities, effectively establishing itself as a reference framework for applying AI techniques to decision support and practical problem-solving.

CRISP-DM continues to be one of the most widely used methodologies for data mining and analytical projects across diverse organisational settings. The framework defines six interconnected and iterative phases: (i) Business Understanding, focused on translating business objectives into analytical goals; (ii) Data Understanding, which addresses data acquisition and initial exploration; (iii) Data Preparation, encompassing data cleaning, integration, and transformation tasks; (iv) Modelling, where statistical and machine learning methods are developed; (v) Evaluation, aimed at assessing model performance and its consistency with business requirements; and (vi) Deployment, in which models are integrated into operational environments to deliver actionable results. The iterative nature of CRISP-DM allows movement between phases based on intermediate findings, ensuring alignment between analytical outputs and organisational objectives. Over time, the framework has been adapted and extended to accommodate a wide range of application domains, including big data analytics, cybersecurity, and financial technologies (Rollins, 2015; Martínez-Plumed et al., 2021; Haakman, 2020).

The increasing convergence of Artificial Intelligence, statistics, data analysis, and computer science, particularly in the context of large-scale data, has led to the consolidation of Data Science as a distinct interdisciplinary field. Concurrently, the widespread availability of Cloud Computing infrastructures has significantly reduced the barriers to developing and deploying data-driven solutions, enabling organisations to scale Data Science initiatives more rapidly. These developments have motivated the emergence of more sophisticated methodological approaches, often proposed and disseminated by major technology providers. Representative examples include the methodologies and workflows promoted by Amazon Web Services (AWS), Microsoft's Team Data Science Process (TDSP), and Google's Data Science frameworks.

In addition to these industry-led initiatives, numerous methodological frameworks have gained recognition within the broader Data Science community. These include CRISP-DM, CRISP-ML(Q), OSEMN, LADM, DDM, Agile Data Science, and SEMMA, among others. From an academic perspective, several structured workflows have also been proposed by researchers such as A. Tandel, J. Thomas, A. Joshi, and P. Guo, along with a growing body of applied research that contributes practical insights into Data Science processes (Díaz et al., 2022; Schmetz et al., 2024; Lonescu et al., 2024; Oakes et al., 2024). This paper reviews both industrial and academic approaches, with particular emphasis on those that demonstrate sufficient methodological rigor, maturity, and applicability to real-world Data Science projects.

Effective Data Science initiatives should be developed within organisations that embrace a data-driven mindset, combining open-standard methodologies such as CRISP-DM with agile analytics principles and established project management frameworks, including PMBOK, during the planning stages. This combination facilitates implementations that are more closely aligned with operational and strategic business requirements. For applications related to maintenance and asset management, compliance with standards such as OSA-CBM (Open System Architecture for Condition-Based Maintenance) is also recommended. In terms of model development, the adoption of workflows and best practices advocated by platforms such as AWS Machine Learning Lens,

Microsoft, and Google is encouraged, alongside established community standards such as CRISP-DM and OSEMN, which have evolved through extensive practical use.

Within this context, the present work adopts a methodology proposed in Moratilla et al. (2023a, 2023b) for conducting data-driven analyses in predictive maintenance scenarios. As illustrated in Table 1, this methodology is organized into five domains comprising a total of 39 processes. By integrating and synthesizing key methodological contributions from both industry and academia, it provides a structured and comprehensive framework specifically designed to address the challenges inherent to predictive maintenance applications.

**Table 1:** Summary of CRISP process.

| # | Domain | Process | Description |
|---|---|---|---|
| 1 | Business Problem | Domain Knowledge | Understanding the business context, constraints, and objectives relevant to the problem. |
| 2 | Business Problem | Data-Driven Approach | Framing the problem so that it can be addressed through data analysis and modelling. |
| 3 | Business Problem | Data Science Approach | Defining the analytical strategy, techniques, and tools to be applied. |
| 4 | Business Problem | Analytics Approach | Selecting appropriate analytical methods to extract insights and support decision-making. |
| 5 | Data Processing | Data Collection | Gathering raw data from relevant sources needed for the analysis. |
| 6 | Data Processing | Data Adequacy | Assessing whether the available data is sufficient, relevant, and representative. |
| 7 | Data Processing | Sampling | Selecting a subset of data that adequately represents the overall dataset. |
| 8 | Data Processing | Data Split (Train, Validation, Test) | Dividing data into subsets to train, validate, and test models. |
| 9 | Data Processing | Data Cleansing | Removing errors, inconsistencies, and noise from the data. |
| 10 | Data Processing | Data Balancing Analysis | Analysing and correcting class imbalance issues in the dataset. |
| 11 | Data Processing | Exploratory Causal Analysis (ECA) | Investigating potential causal relationships among variables. |
| 12 | Data Processing | Exploratory Data Analysis (EDA) | Exploring data patterns, trends, and anomalies through statistical analysis. |
| 13 | Data Processing | Data Visualization | Representing data visually to facilitate understanding and insight generation. |
| 14 | Feature Engineering | Feature Data Transform | Applying mathematical or statistical transformations to features. |
| 15 | Feature Engineering | Feature Importance | Evaluating the relevance and contribution of each feature to the model. |
| 16 | Feature Engineering | Feature Selection | Selecting the most informative features to improve model performance. |
| 17 | Feature Engineering | Feature Extraction | Reducing dimensionality while preserving relevant information. |
| 18 | Feature Engineering | Feature Construction | Creating new features from existing data to enhance predictive power. |
| 19 | Feature Engineering | Feature Transforms | Applying polynomial or nonlinear transformations to features. |
| 20 | Feature Engineering | Feature Learning | Automatically learning feature representations from data. |
| 21 | Model Development | Model Spot Checking | Rapidly testing multiple models to identify promising candidates. |
| 22 | Model Development | Model Evaluation | Assessing model performance using appropriate metrics. |
| 23 | Model Development | Model Selection | Choosing the best-performing model according to predefined criteria. |
| 24 | Model Development | Model Tuning | Optimizing model hyperparameters to improve performance. |
| 25 | Model Development | Model Combination | Combining multiple models to enhance robustness and accuracy. |
| 26 | Model Development | Model Calibration | Adjusting model outputs to better reflect true probabilities. |
| 27 | Model Development | Model Uncertainty Analysis | Analysing uncertainty in model predictions and parameters. |
| 28 | Model Development | Bias-Variance Trade-off Analysis | Evaluating the balance between model complexity and generalization. |
| 29 | Model Development | Model Interpretation | Explaining model behaviour and decision logic. |
| 30 | Model Development | Model Testing | Validating model performance on unseen data. |
| 31 | Model Development | Model Finalization | Preparing the final model for deployment. |
| 32 | Model Development | Model Saving | Storing the trained model for reuse or deployment. |
| 33 | Model Operation | Model Deployment | Integrating the model into a production environment. |
| 34 | Model Operation | Model Execution | Running the deployed model on new data. |
| 35 | Model Operation | Model Analysis | Monitoring model outputs and performance in operation. |
| 36 | Model Operation | Model Updating | Updating or retraining the model as new data becomes available. |
| 37 | AI Systems Audit | AI Framework Audit | Evaluating the overall AI framework and development process. |
| 38 | AI Systems Audit | AI Regulation Audit | Verifying compliance with applicable AI regulations and standards. |
| 39 | AI Systems Audit | AI Models Audit | Auditing models for robustness, fairness, and reliability. |

## III. ANOMALY DETECTION APPROACH

*Unsupervised AE anomaly detection.*

AE-based anomaly detection follows a common blueprint across many condition monitoring domains. First, an AE is trained using only data from healthy operation. At deployment time, the AE reconstructs each new sensor measurement (or set of measurements), and the reconstruction error is computed. If this error exceeds a chosen threshold, the instance is flagged as anomalous (indicating a potential fault). The central design choices in this pipeline include: (i) the architecture of the AE, (ii) the training loss function, and (iii) the strategy for threshold selection.

Early studies in industrial fault detection often used shallow feed-forward AEs (Ahmad et al., 2020), which can model non-linear correlations between variables but treat each time point independently. Subsequent research introduced recurrent architectures (e.g., Long Short Term Memory or Gated Recurrent Unit (GRU) based AEs) to capture temporal dependencies in time-series sensor data (Guo et al., 2018). Temporal models tend to outperform simple "snapshot" models when faults develop gradually over time or manifest as subtle trends, at the cost of increased complexity. Other variants that have been explored include variational autoencoders (VAEs) for probabilistic anomaly scoring (An et al., 2015), convolutional AEs that leverage local structure in high-frequency signals, and even adversarially trained AEs that aim to learn more discriminative latent features. In this study, we focus on a feed-forward dense AE for simplicity and robustness, given our dataset of moderate size and dimensionality.

Recent predictive maintenance literature also emphasises the role of AEs as foundational components within more advanced deep-learning pipelines, including denoising AEs for noise-robust reconstruction, feature-learning AEs used to construct health indicators, and hybrid AE–GAN anomaly detection models that leverage adversarial learning to enhance feature separability (Qian et al., 2022; Yan et al., 2023; Amini et al., 2022). These extensions highlight how AE-based anomaly detection has evolved beyond reconstruction-based scoring into a broader family of deep unsupervised diagnostic tools capable of extracting complex degradation signatures from aero-engine telemetry.

Threshold determination is a crucial part of unsupervised anomaly detection (Hundman et al., 2018). A naive approach is to set a fixed error threshold based on an empirical percentile of the reconstruction error on training data (e.g., flag the worst $p$% of cases as anomalies). However, a static threshold may become suboptimal if the operating regime of the engine changes over time or if sensor noise levels drift. More adaptive schemes have been proposed, such as dynamically adjusting the threshold using moving averages or control charts, or fitting a statistical distribution (e.g., a generalised extreme value distribution) to the tail of reconstruction errors to determine a cutoff with a desired false alarm rate. In our implementation, we experimented with a simple percentile-based threshold as well as a more advanced metric using the Mahalanobis distance. The Mahalanobis approach computes the deviation in a whitened error space, considering the covariance between sensor reconstruction errors. This can improve the detection of subtle faults affecting multiple correlated parameters. By deriving thresholds from the distribution of reconstruction errors on healthy data, our anomaly detector remains fully data-driven and does not require any labelled anomalies.

*Supervised fault classification.*

For comparison, we also consider a traditional supervised learning approach to fault detection (Zhang et al., 2019). Here, the task is treated as a binary classification problem: each time-step (or feature vector) is labelled as either *Normal* (healthy engine operation) or *Fault* (engine experiencing an anomaly or incipient failure). A classifier can then be trained on a labelled dataset to predict this label from the sensor features. In contrast to the AE, a supervised model directly leverages historical fault examples to learn decision boundaries in feature space that distinguish normal vs. faulty patterns.

We included a diverse set of supervised classification algorithms to establish a strong baseline, spanning linear models such as Logistic Regression (LR) and Linear and Quadratic Discriminant Analysis (LDA, QDA) (Anderson, 1958; Duda et al., 1973), probabilistic approaches like Gaussian Naïve Bayes, distance-based models such as k-Nearest Neighbours (k-NN) (Hastie et al., 2017), Support Vector Machines (SVM) known for their robustness in high-dimensional and noisy settings (Cortes et al., 1995; Cristianini et al., 2000), decision trees, and ensemble methods including Random Forests (RFs) (Breiman, 1984; Breiman, 2001), Gradient Boosting, and AdaBoost. We also incorporated neural-network-based methods such as multilayer perceptrons and Artificial Neural Networks (ANNs) (Rosenblatt et al., 1958; Bishop, 1995), which provide flexible non-linear modelling capacity. This diverse collection of models covers a wide range of modelling assumptions and robustness profiles, enabling a comprehensive evaluation of supervised learning performance on helicopter engine telemetry, particularly when fault signatures vary in complexity.

Training a supervised model for fault detection is straightforward if ample labelled data are available: one can minimise a classification loss (e.g., cross-entropy) to fit the training labels. The key challenge, however, is that the performance of such models can degrade if the number of fault examples is very limited or not representative of all possible failure modes. In our case study, we intentionally use the same dataset for both paradigms to illustrate the contrast. The supervised classifiers are given access to the fault labels during training, whereas the AE is not. This way, we can examine how much performance is sacrificed (if any) by not using labels.

## IV. CASE STUDY

To evaluate these approaches, we conducted a case study using a dataset from a helicopter engine monitoring system. The data and experiments are described below.

**DATA DESCRIPTION**

The dataset used in this study originates from the PHM 2024 Data Challenge[1], which provides time-series measurements collected from operational helicopter engines. Each sample represents a high-frequency snapshot of the engine's state, recorded with sub-minute resolution. In total, the dataset comprises approximately 742,625 samples, each containing 7 sensor readings that reflect both external environmental conditions and internal engine dynamics.

The available sensor channels include:
- Outside Air Temperature (OAT) – ambient temperature measured near the engine.
- Mean Gas Temperature (MGT) – average temperature of combustion gases.
- Power Available (PA) – estimated available power based on operating conditions.
- Indicated Airspeed (IAS) – airspeed reported by onboard instrumentation.
- Net Power (NP) – effective mechanical power output of the engine.
- Compressor Speed (CS) – rotational speed of the compressor stage.
- Output Torque (OT) – actual torque delivered by the engine.

Each operational condition corresponds to a known design torque value, which represents the expected torque output for a given configuration. Although this reference value is not explicitly provided, the engine's health can be inferred through the *torque margin*, defined as the difference between actual torque and design torque. A reduction in this margin often indicates performance degradation or incipient failure. This makes torque behaviour, along with associated variables like temperature and speed, a critical indicator in fault detection systems.

Each sample is labelled as either *Normal* or *Anomalous*, based on expert annotations and maintenance records. The *Anomalous* class captures a broad range of abnormal engine behaviours, including transient deviations, performance drops, and early signs of component wear. Importantly, these anomalies do not always correspond to complete engine failures; many are subtle deviations that precede critical deterioration. Approximately 40% of all samples are labelled as anomalous, while the remaining 60% correspond to normal, healthy operation. This relatively high anomaly ratio facilitates the development and evaluation of both supervised and unsupervised fault detection models.

**DATA PREPARATION**

Before modelling, we performed standard data preparation steps to accommodate the different requirements of the supervised and unsupervised approaches. The entire dataset was first split into a fixed test set comprising 10% of all available samples. This hold-out set was used for the final evaluation of both approaches and included a representative mix of normal and anomalous records.

From the remaining 90% of the data, we created separate training partitions for each paradigm:
- Unsupervised approach: Only the normal samples from the 90% subset were retained for training the AE. This ensures that the anomaly detection model learns exclusively from healthy engine behaviour, as required in unsupervised settings. No anomaly labels were used during training or threshold calibration, preventing information leakage.
- Supervised approach: The full 90% subset, comprising both normal and anomalous records, was used to train and validate the classification models. Class labels were available and utilised throughout training, enabling the models to learn direct mappings from sensor features to fault classes.

---

[1] https://data.phmsociety.org/phm2024-conference-data-challenge/

All sensor features were normalised to a common scale using a Min-Max transformation fitted on the training data (normal records only for the unsupervised model, all training data for the supervised case). This rescaling to the [0,1] interval ensures that no single feature dominates due to differing numerical ranges. The same transformation was consistently applied to the test set.

No imputation or outlier removal was necessary, as the dataset was complete and pre-cleaned. Sensor readings were preserved in their original form without smoothing, since short-lived transients or abrupt changes might carry critical information for both detection paradigms.

**EXPLORATORY DATA ANALYSIS**

To gain an initial understanding of the dataset, we conducted an exploratory analysis comparing the behaviour of each sensor feature under normal and anomalous conditions. Figure 1 shows the empirical density distribution of the 7 sensor channels, separated by class. Clear deviations are visible in several variables. For instance, the measured OT (trq_measured) for anomalous samples tends to shift leftward and exhibits greater dispersion, indicating reduced and unstable power delivery. Similarly, features such as OAT, MGT, and PA show distinct distributions between normal and faulty operation, suggesting that anomalies are often accompanied by environmental or performance drift.

Other features like NP and CS show more subtle changes (yet still informative), often narrowing their support or skewing slightly during abnormal conditions. Overall, the density plots reveal that the anomalies are not confined to a single channel but rather manifest across multiple dimensions, which supports the use of multivariate detection methods.

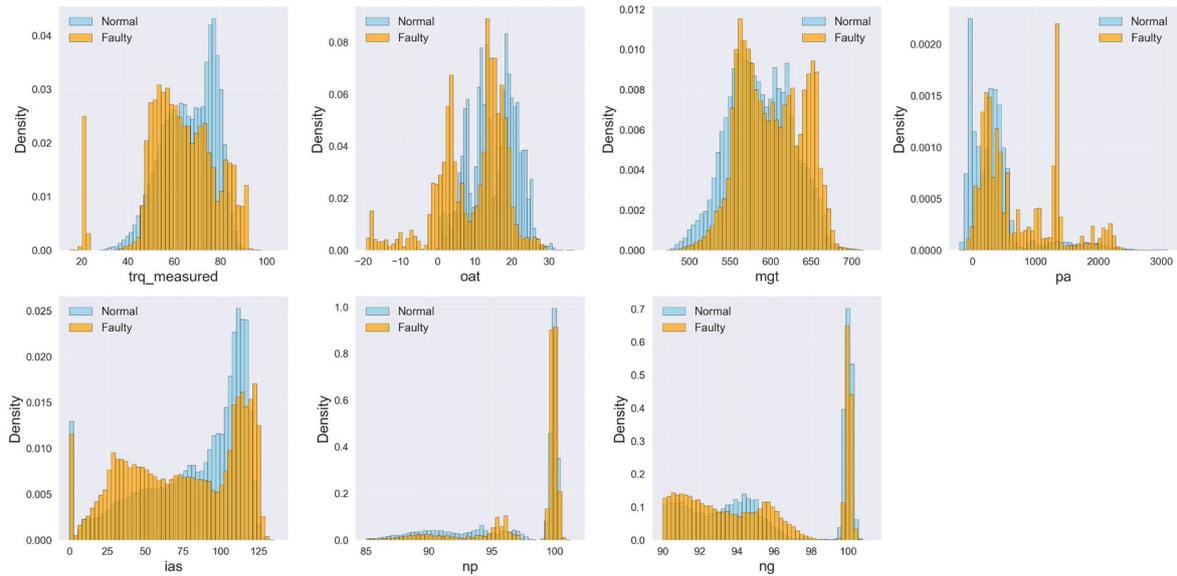

**Figure 1:** Sensor feature distributions under normal and anomalous conditions.

To complement the exploratory analysis, we applied t-distributed Stochastic Neighbour Embedding (t-SNE) to project the 7-dimensional sensor space into three dimensions for visualisation. The resulting embedding, shown in Figure 2, provides a global view of how samples cluster according to operational mode. While there is considerable overlap between normal (green) and anomalous (red) data points, particularly near the central region, many fault samples occupy distinct lobes or boundaries in the embedded space. These structures suggest that at least a subset of faults produces coherent and distinguishable patterns in feature space.

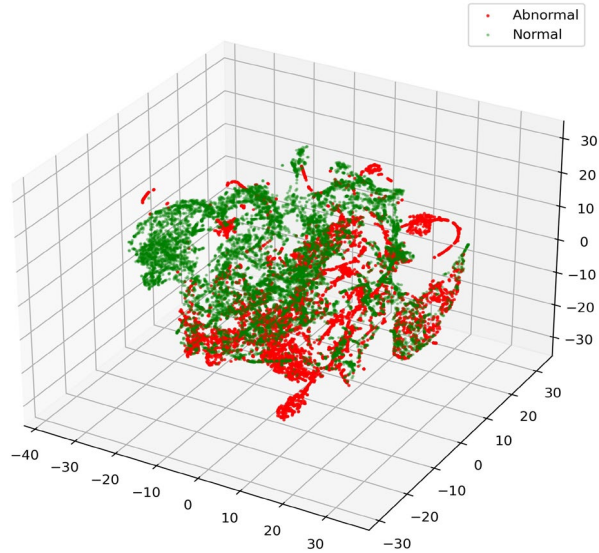

**Figure 2:** Three-dimensional t-SNE embedding of a subset of the helicopter engine dataset.

Together, the distribution plots and t-SNE visualisation suggest that although the boundary between normal and faulty operation is not always sharply defined, there exists sufficient structure in the data to support both supervised and unsupervised learning. The multivariate nature of the deviations also underscores the need for models that can capture non-linear relationships across features.

**FEATURE ENGINEERING**

No complex feature engineering was applied beyond the basic preprocessing steps described earlier (normalisation and dataset partitioning). Since the dataset consists of independent, fixed-length sensor readings rather than continuous time series, there was no need to construct temporal features such as moving averages, lag variables, or spectral descriptors. Each sample is treated as a standalone observation reflecting the engine's instantaneous operating state.

We chose to use the original 7 sensor channels as input features without transformation, under the assumption that they had been selected based on prior domain knowledge to capture the most informative aspects of engine behaviour. These raw features include direct measurements of physical quantities such as temperature, torque, and airspeed, which are known to correlate with engine performance and health.

Although no dimensionality reduction was applied, we verified that all features contributed non-redundant information. None of the channels were constant, and inter-feature correlations remained moderate, indicating that each sensor captured distinct operational signals. Given the small number of features (seven), additional compression was not necessary. This minimalist approach allows the models, especially the AE, to learn data-driven latent representations (González-Muñiz et al., 2022) directly from the raw sensor inputs, without relying on handcrafted transformations or domain-specific feature extraction pipelines.

By preserving the native structure of the data, we ensure a fair comparison between supervised and unsupervised models, and we test their ability to infer health indicators directly from raw engine measurements.

**MODEL**

For the supervised learning approach, we considered a suite of binary classification models as described in Section III. For this approach, the classifiers included: LR (a linear model), Gaussian Naïve Bayes, SVM with an RBF kernel, k-NN, Decision Tree, RF, Gradient Boosting Machine, AdaBoost (with a decision tree base estimator), and a multilayer perceptron (feed-forward neural network). These models were implemented using scikit-learn and chosen to span a range from simple (e.g., linear, instance-based) to complex (e.g., ensemble, neural network) approaches commonly used in fault diagnosis. The MLP classifier was configured with a relatively small architecture (a single hidden layer) to reduce the risk of overfitting given our dataset size.

On the unsupervised side, we designed a feed-forward AE to perform anomaly detection. The AE has a symmetric architecture with an encoder that compresses the input features into a low-dimensional latent representation, and a decoder that reconstructs the input from the latent code. In particular, the network structure is 7-5-3-5-7: it takes the 7-dimensional sensor vector as input, then a dense hidden layer of 5 neurons (with exponential linear unit (ELU) activation) reduces the dimensionality, followed by a bottleneck layer of 3 neurons. The decoder then expands from the 3-dimensional code to 5 neurons (ELU) and finally back to 7 output neurons. This architecture was chosen to force a significant compression (down to 3 features) of the normal data, ensuring that the model captures the most salient features of healthy engine behaviour. By comparison, using a latent size equal to the input (7) would risk the AE simply learning the identity function without extracting useful features, whereas a very small latent size (e.g. 1-2) proved inadequate during preliminary testing. The capacity of the 7-5-3-5-7 network was found to be sufficient to reconstruct normal operating data with low error while still challenging the model to generalise.

**TRAINING PROTOCOL**

After initialising the models as above, we proceeded to train them using the prepared dataset. For the supervised models, training involved providing the labelled *Normal/Anomalous* samples from the training set and optimising each model's objective (e.g., maximising likelihood or minimising classification error). We used 5-fold stratified cross-validation on the training set to guide model selection and hyperparameter tuning. Model performance was primarily measured by the F1-score (the harmonic mean of precision and recall for the Anomalous class) during validation, since F1 offers a balance between catching as many anomalies as possible (recall) and avoiding false alarms (precision). Hyperparameters for certain models were tuned based on cross-validation: for example, the number of neighbours k in k-NN and the regularisation strength in LR were adjusted to maximise F1. Ensemble models like the RF and Gradient Boosting were run with a sufficient number of estimators (100 trees) to ensure stability, and default values were used for parameters like tree depth unless cross-validation indicated otherwise. Each final classifier was then retrained on the entire training set (with the chosen hyperparameters) before evaluating on the test set.

The AE was trained using only the *Normal* class training data (approximately 354,566 samples). Its training objective was to minimise the mean squared error (MSE) between the input and the reconstructed output. We used the Adam optimiser with an initial learning rate of 0.001. Training was done in mini-batches of size 1,024 for efficiency. To prevent overfitting and determine an optimal stopping point, we employed an early stopping strategy: 10% of the normal training data was held out as a validation set for the AE, and the model's validation loss was monitored. If the validation reconstruction error did not improve for 25 epochs, training was stopped and the best weights were restored. We also implemented a learning rate scheduler (ReduceLROnPlateau) that reduced the learning rate by a factor of 0.2 if the validation loss plateaued for 20 epochs (with a minimum learning rate of $1\times10^{-6}$). The AE was trained for a maximum of 200 epochs, but in practice, it converged much earlier (roughly within 50 epochs) as the validation loss stabilised. At convergence, the AE reliably reproduced healthy engine sensor patterns with a very low average error, forming the basis for our anomaly detection mechanism.

**MODEL ADAPTATION**

Certain adaptations were applied to tailor the models to the helicopter engine scenario. For the AE, one important consideration was the threshold for flagging anomalies from the reconstruction error. Rather than fixing an arbitrary threshold, we derived it empirically from the training reconstruction error distribution. We computed the MSE for every training sample (all of which are healthy) after training the AE. We then set the anomaly threshold as a high percentile of this error distribution. In our experiments, we found that using the $85^{th}$ percentile of the training error distribution worked well: this meant that about 15% of the healthy training points would lie above the threshold (to be conservative, simulating an acceptable false alarm rate on training data), and anything with error higher than that was considered abnormal. We also evaluated a thresholding approach based on the Mahalanobis distance (Lin et al., 2010). To do this, we calculated the covariance matrix of the reconstruction residuals on the training set and used its inverse to compute a Mahalanobis distance for each sample's error vector. This effectively accounts for the correlations between features in how we measure "distance" from the normal manifold. An $85^{th}$ percentile threshold was likewise applied to the Mahalanobis distances of the training set to classify anomalies. We observed that the Mahalanobis-based threshold yielded slightly better results (higher recall for a given precision) than using raw MSE alone, confirming that considering error covariance is beneficial in our case. Unless otherwise noted, results reported for the AE use the Mahalanobis-distance method for determining anomalies.

Another adaptation relates to the dynamic nature of engine data. Engine operating conditions (idle, cruise, takeoff, etc.) can cause legitimate shifts in sensor readings. We did not explicitly segment the data by operating mode in this study; however, in a deployed system, one might incorporate the engine's operating regime into the anomaly detection logic (for instance, by having separate thresholds for different modes or by feeding the mode as an input to the model). In our dataset, the t-SNE analysis (Li et al., 2024) and model performance suggested that the features were effective at capturing anomalies without needing mode-specific handling, but this remains a consideration for future work.

For the supervised models, the main problem-specific concern was the class imbalance, albeit in our dataset the imbalance was not severe (roughly 3:2). We still took care to use performance metrics (like F1-score and area under the ROC curve) that account for imbalanced classes rather than relying solely on accuracy. During cross-validation, we also ensured that folds were stratified so that each fold had a representative class balance. No custom cost weights were applied to the classes given the relatively balanced nature of the data; in scenarios with more extreme imbalance (e.g., 1% anomalies), one might introduce weighting or oversampling strategies to ensure the classifier pays sufficient attention to the minority class.

## V. EVALUATION

To assess and compare the performance of the supervised and unsupervised approaches, we established a consistent evaluation protocol on the test set. Each model (classifier or AE) produced a binary decision for each test sample, indicating whether that sample was predicted to be *Normal* or *Anomalous*. For the supervised classifiers, this decision was directly the model's output (using a 0.5 probability threshold for models like LR or the class prediction for non-probabilistic models like SVM and k-NN). For the AE, the decision was obtained by computing the reconstruction error for each test sample and comparing it to the anomaly threshold determined during training (as described in the Case Study section). Any test sample with error above the threshold was classified as *Anomalous* by the AE; otherwise, it was considered *Normal*.

We evaluated the predictions using several standard metrics: accuracy (overall proportion of correct predictions), precision for the anomalous class (also called positive predictive value, the fraction of predicted anomalies that were true anomalies), recall for the anomalous class (also known as sensitivity or true positive rate, the fraction of true anomalies that were correctly detected), and the F1-score for the anomalous class (the harmonic mean of precision and recall). We also calculated the Area Under the Receiver Operating Characteristic Curve (AUROC) where applicable, to summarise the trade-off between true positive rate and false positive rate across different thresholds. For the AE, we can obtain an AUROC by varying the anomaly threshold from very low (lenient, flagging almost everything as anomaly) to very high (strict, flagging almost nothing) and comparing against the ground truth labels. However, since we fixed a threshold for deployment, our primary focus was on the precision, recall, and F1 at that operating point.

It is important to note how class imbalance can affect these metrics. In our test set, roughly 40% of instances are anomalies (which is higher than typical in many real-world scenarios). This means accuracy alone can be a misleading indicator of performance, detecting all normals correctly but missing anomalies could still yield a high accuracy due to the large number of normal samples. Therefore, our analysis emphasises precision, recall, and F1 for the anomaly class, which more directly reflect the fault detection capability.

The evaluation protocol was as follows: each model was applied to the entire test set to produce predictions. No further training or threshold tuning was done on the test data (to avoid bias). We then computed the aforementioned metrics by comparing predictions to the true labels. In addition to numerical metrics, we also inspected confusion matrices and error distributions to qualitatively understand the mistakes each method made (e.g., what kinds of anomalies tended to be missed by the AE versus those missed by the classifiers).

## VI. RESULTS

Table 2 summarises the detection performance of the leading supervised model versus the AE approach on the test set. The RF emerged as the top-performing supervised classifier, achieving an almost perfect score across metrics. It attained a recall of 99.95% on anomalies, meaning it missed virtually none of the faults in the test data, and a precision of 99.93%, indicating almost no false alarms, with an F1-score above 0.999. In fact, several of the supervised models (including Gradient Boosting and even k-NN) reached very high performance, reflecting that the anomalies in this dataset are highly separable from normal data when the models are trained with labelled examples. At the lower end of the supervised spectrum, the LR classifier achieved about 90% F1, and Naïve Bayes was substantially worse (around 65% F1), likely due to its simplistic assumptions. The MLP neural network, after some tuning, also performed well (around 99.8% F1), although an initial configuration without tuning had struggled, underscoring the importance of hyperparameter choices.

Table 2: Performance comparison of a top-performing supervised classifier (RF) and the unsupervised AE on the helicopter engine test set.

| Model | Precision | Recall | F1-score | Accuracy |
|---|---|---|---|---|
| Random Forest (Supervised) | 0.9993 | 0.9995 | 0.9994 | 0.9995 |
| AEs (Unsupervised) | 0.8181 | 0.8856 | 0.8505 | 0.8758 |

The AE-based anomaly detector, using the Mahalanobis-distance threshold, achieved a precision of about 82% and a recall of roughly 89% on the test data. In practical terms, this means that when the AE flagged an engine state as anomalous, 82 out of 100 times it was correct (a true fault), and it successfully detected about 89 out of every 100 true anomaly instances. The F1-score for the AE was 0.85, which, while lower than the near-perfect supervised model, is still indicative of good performance given no fault examples were used in training. The accuracy of the AE's decisions was around 88%. Most of the errors made by the AE came from false positives rather than missed anomalies. With a recall of 88.56%, the model successfully detected the vast majority of anomalous samples, missing only about 11.4% of true faults. However, its precision was lower—at 81.81%—indicating that nearly 18% of the samples flagged as anomalies were in fact false alarms. This reflects a more aggressive detection strategy: the model errs on the side of caution, preferring to raise an alert even at the risk of flagging some normal instances. Such behaviour is typical in unsupervised settings where no labelled faults are seen during training, and the anomaly threshold must be derived heuristically. While this can increase the operational burden due to unnecessary inspections, it helps ensure that subtle or early-stage faults are less likely to be missed.

Comparing the two approaches, the supervised RF clearly had an edge in this scenario, as it could almost perfectly separate the two classes given the wealth of labelled anomaly data. The unsupervised AE, however, proved capable of catching the majority of anomalies with minimal false alarms, despite never seeing an anomaly during training. These results highlight that an AE is a viable solution for fault detection, especially in situations where labelling anomalies is infeasible. If one has the luxury of many labelled fault examples (as we did in this dataset), a supervised classifier will likely outperform an unsupervised method. But if such labels are unavailable, the AE still delivers strong results and can serve as an effective early warning system.

To further explore the AE's behaviour, we examined a few specific instances from the test set. In general, the anomalies that the AE missed tended to be those where the deviation from normal was relatively small or affected a sensor that the AE did not heavily weight in its reconstruction error. For example, a mild anomaly that only caused a slight increase in one temperature sensor might slip under the threshold. On the other hand, the anomalies it did detect were typically more pronounced or multidimensional (affecting several sensors at once), which produced a spike in reconstruction error that easily exceeded the threshold. In contrast, the misclassifications made by the RF were almost nonexistent; it occasionally gave a false alarm on a borderline normal sample that had an unusual combination of sensor readings, but such cases were extremely rare.

## VII. CONCLUSIONS AND FUTURE WORK

*Limitations and future work.*

While the results are encouraging, several limitations in our current study suggest avenues for future work. First, the dataset used contains an unusually high proportion of labelled anomalies (approximately 40%), although it is derived from a real helicopter engine scenario. In operational contexts, faults are typically rare events, often comprising less than 5% or even less than 1% of recorded data. The abundance of anomalies in this dataset benefits supervised classifiers by providing ample fault examples for training, but it does not reflect the class imbalance seen in real-world deployments. Future evaluations should include more realistic, imbalanced datasets to assess whether the conclusions hold under more typical conditions, and to test the robustness of each model against the scarcity of fault labels.

Second, our unsupervised approach relied on a relatively simple AE architecture with a fixed anomaly threshold derived from the training set. While this setup proved effective for our static, snapshot-based data, it lacks adaptability to changes in operational context. Since the dataset consists of independent observations and does not capture time evolution, the model cannot leverage temporal continuity to detect trends or progressive degradation. However, in real applications, even non-temporal data streams may experience slow drifts due to component ageing, sensor recalibration, or environmental variation. Without mechanisms to periodically update the model or recalibrate thresholds, the detector may accumulate false positives or start overlooking new types of faults. Future work could explore adaptive learning strategies such as incremental retraining with fresh normal data, online threshold adjustment, or anomaly feedback integration.

Third, the scope of this study was limited to binary anomaly detection, i.e., deciding whether a given data point is normal or not. While useful for early fault detection, this approach does not offer diagnostic insights (e.g., identifying the type or cause of failure) or prognostic forecasts (e.g., estimating RUL). Extending the supervised approach to multi-class classification could allow fault-type identification if labelled categories become available. Likewise, incorporating regression models to predict degradation severity or time-to-failure would be valuable for more proactive maintenance planning. A hybrid framework that integrates unsupervised anomaly flagging with downstream diagnosis or prognostics could offer a more comprehensive solution.

Finally, model explainability remains a key challenge, especially in safety-critical domains such as aerospace. Although the RF classifier offers partial transparency via feature importance rankings, its decision process may still be opaque to domain experts. The AE is even more difficult to interpret, as its anomaly scores emerge from latent representations with no direct semantic meaning. Future research should consider interpretable anomaly scoring methods. Enhancing interpretability would improve user trust and facilitate actionable maintenance decisions when an alert is raised.

*Conclusions.*

This study evaluated the viability of unsupervised learning for fault detection in helicopter engines by comparing an AE-based anomaly detector trained solely on healthy data, with a fully supervised classifier trained on labelled faults. While the supervised model (a RF) achieved near-perfect precision and recall, it required access to a large, balanced dataset that included a significant number of annotated failure instances. In contrast, the unsupervised AE attained an F1-score of 0.85, with reasonably high recall (88.6%) and precision (81.8%), despite having no exposure to faults during training.

These results highlight the core trade-off between the two paradigms. Supervised models can deliver superior accuracy but depend on a rich dataset of historical faults, which may not be available for many real-world systems, especially in high-reliability domains like aerospace. Conversely, the unsupervised approach offers a practical, label-free alternative that remains effective in identifying a substantial portion of anomalies. Its performance, while lower, is still robust enough to support early-warning applications, particularly in scenarios where undetected degradation poses operational risks.

Importantly, the unsupervised method does not require any prior knowledge of what faults look like. This makes it especially valuable in contexts where failure modes are poorly understood, rare, or evolving over time. The ability to deploy a detection system trained exclusively on normal behaviour reduces the burden of data curation and allows organisations to begin monitoring immediately. As additional operational data accumulate, the unsupervised model can help identify previously unseen patterns and build the foundation for more advanced, supervised solutions.

Our findings demonstrate that unsupervised AEs are a viable and deployable option for predictive maintenance in helicopter engines, even if they do not reach the precision of supervised models trained on fault data. Their ease of implementation, low data requirements, and generalisation capability make them a compelling choice for real-world applications where labelled faults are scarce. In the long term, a hybrid approach may offer the best of both worlds: AEs provide continuous monitoring and anomaly flagging, while supervised classifiers refine diagnosis as labelled fault data become available. This strategy enables gradual improvement of system reliability and paves the way for more intelligent, adaptive maintenance pipelines in the aerospace industry.

# ACKNOWLEDGEMENTS

This work has been carried out within the framework of the "Cátedra ENIA IA[3]: Cátedra de Inteligencia Artificial en Aeronáutica y Aeroespacio", subsidised by the "Ministerio de Asuntos Económicos y Transformación Digital" (Secretaría de Estado de Digitalización e Inteligencia Artificial), del Gobierno de España.